%% file: main.tex
\documentclass[conference]{IEEEtran}
\IEEEoverridecommandlockouts
\usepackage{cite}
\usepackage{amsmath,amssymb,amsfonts}
\usepackage{algorithmic}
\usepackage{graphicx}
\usepackage{textcomp}
\usepackage{xcolor}
\usepackage[utf8x]{inputenc}
\usepackage{multirow}
\usepackage{adjustbox}
\usepackage[absolute]{textpos}

\usepackage[acronym]{glossaries}
\newacronym{SOD}{SOD}{Salient Object Detection}

\def\BibTeX{{\rm B\kern-.05em{\sc i\kern-.025em b}\kern-.08em
    T\kern-.1667em\lower.7ex\hbox{E}\kern-.125emX}}
\begin{document}

\begin{textblock*}{\textwidth}(1.5cm,0.2cm)
\small{\copyright~2023 IEEE. Personal use of this material is permitted. Permission from IEEE must be obtained for all other uses, in any current or future
media, including reprinting/republishing this material for advertising or promotional purposes, creating new collective works, for resale or
redistribution to servers or lists, or reuse of any copyrighted component of this work in other works.\\
This file corresponds to the accepted version of the manuscript published in the 2023 Congress in Computer Science, Computer Engineering,
\& Applied Computing (CSCE), Las Vegas, NV, USA, 2023, pp. 1960-1968, doi: 10.1109/CSCE60160.2023.00323}
\end{textblock*}

\title{The Impact of Background Removal on Performance of Neural Networks for Fashion Image Classification and Segmentation}

\author{\IEEEauthorblockN{Junhui Liang}
\IEEEauthorblockA{\textit{Department of Computer Science} \\
\textit{KTH Royal Institute of Technology}\\
Stockholm, Sweden \\
junhuil@kth.se}
\and
\IEEEauthorblockN{Ying Liu}
\IEEEauthorblockA{\textit{Norna} \\
Stockholm, Sweden \\
ying.liu@norna.ai}
\and
\IEEEauthorblockN{Vladimir Vlassov}
\IEEEauthorblockA{\textit{Department of Computer Science} \\
\textit{KTH Royal Institute of Technology}\\
Stockholm, Sweden \\
vladv@kth.se}
}

\maketitle

\begin{abstract}
Fashion understanding is a hot topic in computer vision, with many applications having great business value in the market. Fashion understanding remains a difficult challenge for computer vision due to the immense diversity of garments and various scenes and backgrounds. In this work, we try removing the background from fashion images to boost data quality and increase model performance. Having fashion images of evident persons in fully visible garments, we can utilize Salient Object Detection to achieve the background removal of fashion data to our expectations. A fashion image with the background removed is claimed as the "rembg" image, contrasting with the original one in the fashion dataset. We conducted extensive comparative experiments with these two types of images on multiple aspects of model training, including model architectures, model initialization, compatibility with other training tricks and data augmentations, and target task types. Our experiments show that background removal can effectively work for fashion data in simple and shallow networks that are not susceptible to overfitting. It can improve model accuracy by up to 5\% in the classification on the FashionStyle14 dataset when training models from scratch. However, background removal does not perform well in deep neural networks due to incompatibility with other regularization techniques like batch normalization, pre-trained initialization, and data augmentations introducing randomness. The loss of background pixels invalidates many existing training tricks in the model training, adding the risk of overfitting for deep models.
\end{abstract}


\section{Introduction}
\label{sec:introduction}
\input{sections/introduction}

\%input{sections/preliminaries}

\section{Controlled Comparative Experiments Design}
\label{sec:experiments design}
\input{sections/experiments_design}

\section{Evaluation of the Impact of Background Removal on Fashion Image Classification}
\label{sec:image classificaion}
\input{sections/image_classificaion}

\section{Evaluation of the Impact of Background Removal on Fashion Image Segmentation}
\label{sec:image segmentation}
\input{sections/image_segmentation}

\section{Conclusions}
\label{sec:conclusions}
\input{sections/conclusions}

\bibliographystyle{IEEEtran}
\typeout{}
\bibliography{references}
\end{document}

%% file: sections/introduction.tex
Fashion analysis has been a favored domain in computer vision thanks to its great
business value in the online shopping experience. With the progress of computer vision and deep learning, there are many outstanding applications in image retrieval, product recognition, style recommendation, and competitor analysis in the fashion market. Most fashion datasets consist of images with clothes visible in various scenes such as online shopping, daily life, celebrity event, etc. In addition to diverse garments, various backgrounds make fashion data challenging for automated fashion analysis using computer vision methods.

This work evaluates whether background removal can augment data quality and improve model performance on fashion data. Inspired by data augmentation and attention mechanism, background removal removes pixels of specific regions to filter noises at the data level while maintaining the entire fashion object visible. It is expected to increase model accuracy at the expense of speed. To remove the background of fashion images, we apply \gls{SOD}~\cite{wang_salient_2021}, which aims to segment the most visually attractive objects and helps to remove the background of fashion images to our expectations cleanly. To evaluate and validate the value of background removal for fashion data analysis using machine learning, we conduct comparative experiments on various model architectures with the original images and images without background that we call \emph{rembg} images. 

More specifically, we focus on the following questions. Is it possible to remove the background of fashion images cleanly? Can background removal boost fashion data quality? What situation and aspects of model training can background removal positively affect model performance? How about the compatibility of background removal with other data augmentations or training tricks? Is background removal necessary for fashion data if only concerning the model accuracy? 

To comprehensively understand the impact of background removal on model performance, we design and conduct extensive experiments to compare two different inputs, namely the original and \emph{rembg} images, in the following aspects of model training: various model architectures such as backbone, network depth, normalization layer, various initialization consisting of random initialization and pre-trained initialization, compatibility with other data augmentation, and various task types including classification on FashionStyle14~\cite{takagi_what_2017}, instance segmentation and semantic segmentation on Fashionpedia~\cite{jia_fashionpedia_2020}.

With extensive experiments on background removal, we found that background removal is unnecessary for fashion data in most situations. Background removal can eliminate background interference and assist the model in focusing on key regions. However, due to the loss of background information, background removal greatly increases the risk of overfitting deep networks. It weakens the ubiquitous regularization techniques that can release overfitting, such as batch normalization, pre-trained initialization based on transfer learning, and other common data augmentation extending datasets with similar data. The incompatibility with existing training measures and tricks makes background removal only play a positive role in simple and shallow networks, which are not susceptible to overfitting. Moreover, background removal can merely work in the classification task, while instance and semantic segmentation can still provide location annotations unaffected by background removal. 

In summary, our results indicate that background removal is a good choice for  shallow networks trained from scratch in the classification task but not a necessary option for fashion data if using a deep network or pre-trained initialization. 

Our contributions in this paper are summarized as follows.
\begin{itemize}
    \item We empirically study the impact of background removal on the performance of neural network models for fashion image analysis. We obtain \emph{rembg} images (images without background) using a Salient Object Detection tool and compare the performance of models trained on different inputs, original or \emph{rembg} images of multiple datasets.
    \item We testify background removal from fashion images in various model architectures such as network depth, normalization layers, and backbone.
    \item We verify the background removal from fashion images in different model initializations, including random and pre-trained initializations.
    \item We examine background removal's compatibility with other data augmentations like rotate, flip, etc.
    \item We validate background removal from fashion images in various task types, including classification, instance segmentation, and semantic segmentation.
\end{itemize}

%% file: sections/experiments_design.tex
In this section, we present the design of our controlled comparative experiments to verify the effect of background removal from fashion images on the accuracy of fashion image classification and segmentation with different model training methods and parameters. 

To comprehensively research the value of background removal for fashion understanding in every aspect of model training, we design the controlled experiments respectively on model architecture, model initialization, data augmentation, and task types, including classification, instance segmentation, and semantic segmentation. More specifically, we design and perform several groups of controlled comparative evaluation experiments for the following model aspects. 
\begin{itemize}
    \item \textbf{Model architecture.} First, we testify to the effect of background removal on the performance of different model architectures with various key architecture attributes, including network depths, normalization layers, backbones, etc.
    \item \textbf{Model initialization.} Second, we verify the value of background removal on the performance of the models with various model initialization, often categorized into two kinds: (1) training a model from scratch with random initialization or (2) fine-tuning a model from the one pre-trained on ImageNet.~\cite{deng_imagenet_nodate}. 
    \item \textbf{Data augmentation.} Third, we evaluate the compatibility of background removal, 
    considered as one kind of data preprocessing, with other data preprocessing techniques, namely, data augmentation techniques used for images to boost model performance and generalization, such as crop, flip, rotate, shear, brightness, Solarize, etc.
    \item \textbf{Task Type.} Lastly, we evaluate the effect of background removal on the performance of different ML tasks on fashion data, including classification, instance segmentation, and semantic segmentation.
\end{itemize}

By evaluating the effect of background removal on the performance of fashion image classification models for the above aspects, we can more comprehensively learn about the significance of background removal for fashion data and decide whether background removal is a necessary preprocessing for fashion data to improve fashion model performance.

\subsection{Neural Networks Used in Evaluation Experiments}
\label{subsec:neural networks }
IN this study, we evaluate the impact of background removal on the performance of Convolutional Neural Networks (CNNs) for image classification and segmentation. CNN~\cite{dumoulin_guide_2018} is a widely used Artificial Neural Network for image processing and understanding. It has become the mainstream approach in computer vision, applied in various tasks such as image classification, object detection, and image segmentation. The core components of CNN are the convolutional, pooling, and fully connected layers. The convolutional layer extracts features from images using learnable kernels, while the pooling layer down-samples the feature maps, reducing model parameters and enabling efficient training. The fully connected layers serve as classifiers for output prediction based on the extracted features. 
For our experiments, we selected VGG11, VGG13, VGG16, and VGG19~\cite{simonyan_very_2015} as representative of shallow and simple models, ResNet-50~\cite{he_deep_2015}, ResNet-101~\cite{he_deep_2015} as typical deep networks, and ResNeSt-101~\cite{zhang_resnest_2020} as a deep network with attention architecture. For image instance segmentation, we used ResNeXt-101 and ResNeSt-101 to assess the impact of background removal on the model's performance.

\subsection{Data Sets}\label{subsec:data sets}
 To evaluate the benefit (if any) of background removal for fashion image classification, we used the FashionStyle14 dataset~\cite{takagi_what_2017}, where each image represents a fashion style and has visible and salient fashion objects. To evaluate the effect of background removal on the performance of instance segmentation and semantic segmentation, we used Fashionpedia dataset~\cite{jia_fashionpedia_2020}, in which, on average, each image shows one person with several main garments, accessories, and garment parts, each annotated by a tight segmentation mask. 

 \subsection{Background Removal}\label{subsec:background removal}
\begin{figure}[ht]
    \centering
    \includegraphics[width=\columnwidth]{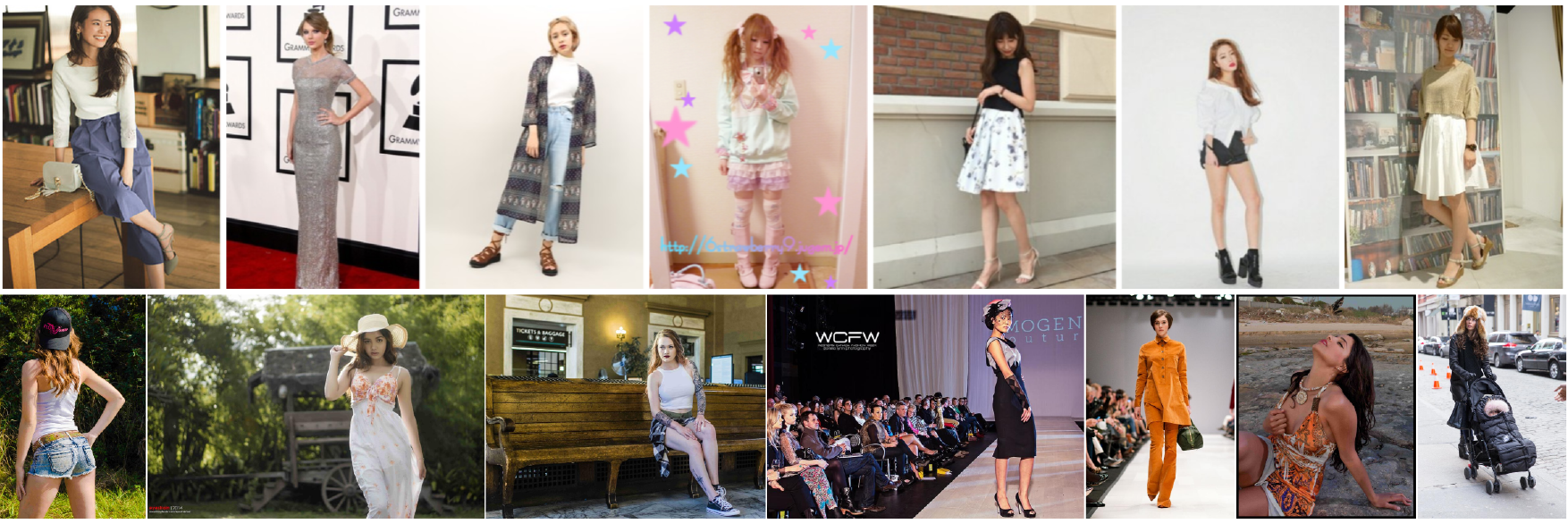}
    \caption{Original image samples from FashionStyle14~\cite{takagi_what_2017} and Fashionpedia~\cite{jia_fashionpedia_2020}.}
    \label{fig:sample-of-images}
\end{figure}

Inspired by data augmentation and attention mechanisms, background removal is to remain the entire fashion object visible and remove pixels of specific regions to filter noises at the data level. 
A typical fashion image, such as those in the FashionStyle14~\cite{takagi_what_2017} and Fashionpedia~\cite{jia_fashionpedia_2020} datasets, is a clothing image in daily life, street style, celebrity events, runway, and online shopping annotated by fashion experts. Figure~\ref{fig:sample-of-images} shows several sample images from FashionStyle14 and Fashionpedia datasets used in this study. The common feature of most fashion images is having visible key objects of fashion coordinate, i.e., usually, one or several evident salient persons in assorted garments in the center of an image, which is easily distinguished from diverse backgrounds. This feature of fashion images to have salient persons in clothes perfectly matches the concept of Salient Object Detection (\gls{SOD})~\cite{wang_salient_2021} aiming to segment the most visually attractive objects in an image. Naturally, one can use a SOD method, e.g., $U^2$-Net~\cite{qin_u2-net_2020}, to remove the background of fashion data. For our comparative evaluation experiments on fashion images, we chose the Rembg~\cite{gatis_rembg_2022} tool based on $U^2$-Net to remove the background 
 while keeping salient persons in clothes intact. 

\begin{figure}[ht]
    \centering
    \includegraphics[width=\columnwidth]{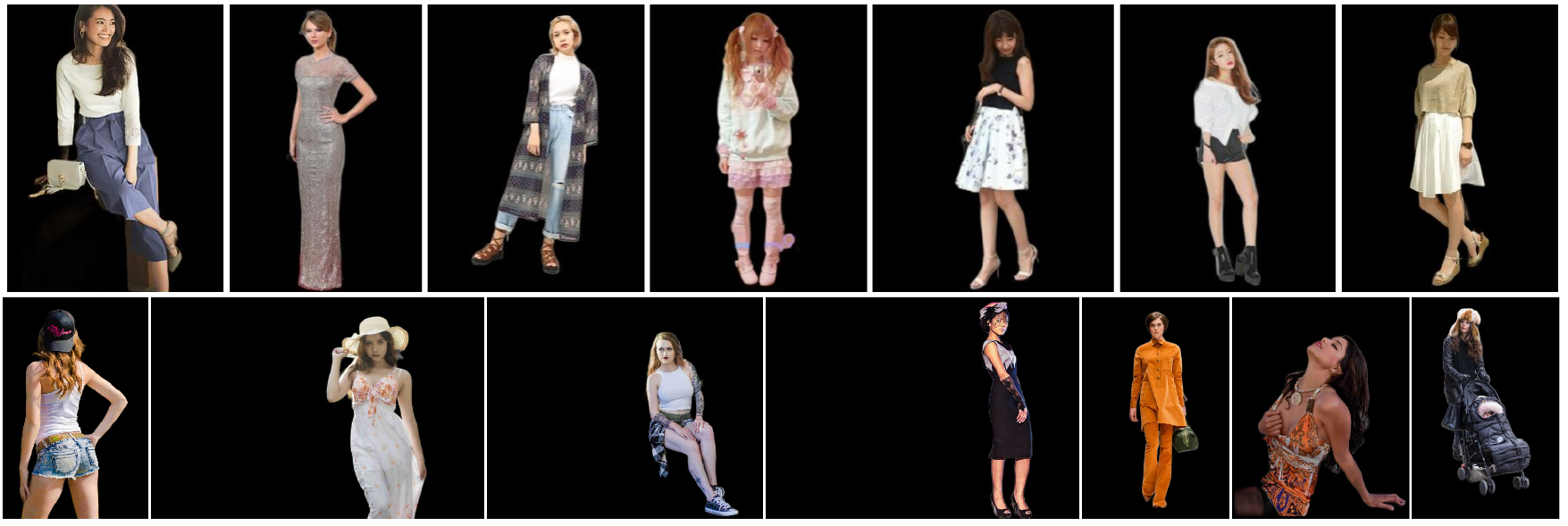}
    \caption{\emph{Rembg} image samples from FashionStyle14~\cite{takagi_what_2017} and Fashionepdia~\cite{jia_fashionpedia_2020}.}
    \label{fig:sample-of-rembg-images}
\end{figure}

To validate the value of background removal for fashion data, we conduct our comparative evaluation experiments with the original images and the corresponding images without background, called \emph{rembg} images, obtained from the original images using the Rembg tool. Figure~\ref{fig:sample-of-rembg-images} illustrates the effect SOD applied to fashion data to remove background, which is ideal, basically meeting the requirements of background removal we expected. 

It is worth noting that some fashion images might not be feasible to remove the background using SOD, such as the images of product details and the images containing half or part of models in garments. 
When the key objects are not fully displayed and not placed in the center of the image, the SOD method might destroy the original images losing information and garments rather than purely removing the image background. Furthermore, the black clothes with high contrast to the rest of the image are easily ignored by SOD. Fortunately, the datasets chosen for our experiments are pre-processed by manual selection of sensible criteria so that each image has visible and salient fashion objects. 

\subsection{The General Pipeline for Comparative Experiments}\label{subsec:experimental pipelines}
\begin{figure}[ht]
    \centering
    \includegraphics[width=1\columnwidth]{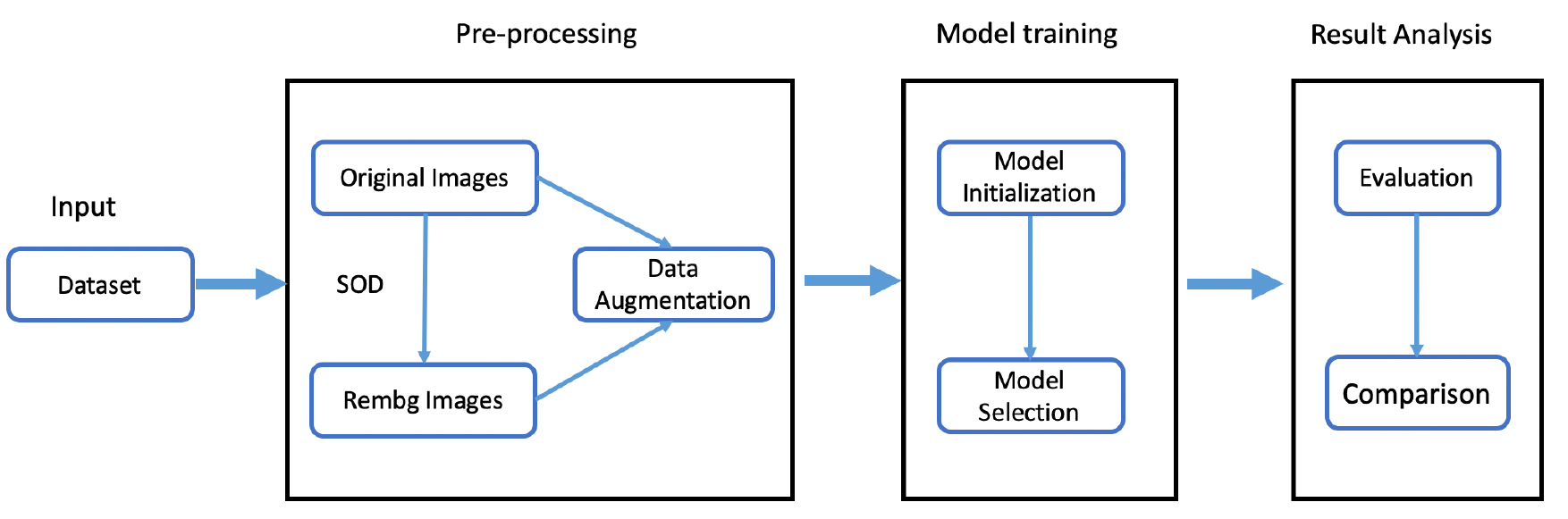}
    \caption{The general pipeline of comparative experiments.}
    \label{fig:pipeline-of-experiment}
\end{figure}

Our general pipeline for controlled comparative experiments shown in Figure~\ref{fig:pipeline-of-experiment} comprises three stages: preprocessing, model training, and result analysis. To comprehensively research the value of background removal for fashion data, we control experiments through the whole procedure of 
ML model training. We compare the effect of background removal in image classification and segmentation for different inputs in various models and training settings. The choice of input fashion dataset depends on the targeted task and its expected output. As mentioned in Section~\ref{subsec:data sets}, we choose FashionStyle14~\cite{takagi_what_2017} for image classification and Fashionpedia~\cite{jia_fashionpedia_2020} for instance and semantic segmentation to validate the benefit of background removal in different fashion tasks. The preprocessing stage includes background removal and data augmentation. We remove the background from the fashion images using the SOD tool~\cite{gatis_rembg_2022}. The fashion images without backgrounds are marked with \emph{rembg}. Data augmentation for images, e.g., crop, flip, rotate, shear, brightness, Solarize, etc., is also one kind of preprocessing. The model training stage contains model initialization, such as random and pre-trained initialization, model selection for various model architectures, and backbones. Finally, we evaluate model performances on a specific task by specific evaluation metrics to verify whether background removal can enhance model performance under the same settings as the original images.

\subsection{Experiment Settings}\label{subsec:experiment settings}
We conducted experiments on the following two personal computers: Core\textsuperscript{\texttrademark} i9-9900KF and  Intel\textsuperscript{\textregistered} Core\textsuperscript{\texttrademark} i7-8700K, each  with GPU Nvidia GeForce RTX 2080 Ti.
All our model codes are based on the open-source toolboxes of OpenMMLab~\cite{openmmlab_2020}, including MMClassification~\cite{mmclassification_contributors_openmmlabs_2020}, MMDetection~\cite{mmdetection_contributors_openmmlab_2018}, and MMSegmentation~\cite{mmsegmentation_contributors_openmmlab_2020}.

%% file: sections/image_classificaion.tex
First, we present our evaluation of the impact of background removal on the performance of representative Neural Networks for image classification on the FashionStyle14~ \cite{takagi_what_2017} dataset.

\subsection{The pipeline for image classification experiments.} 
\begin{figure}[ht]
    \centering
    \includegraphics[width=\columnwidth]{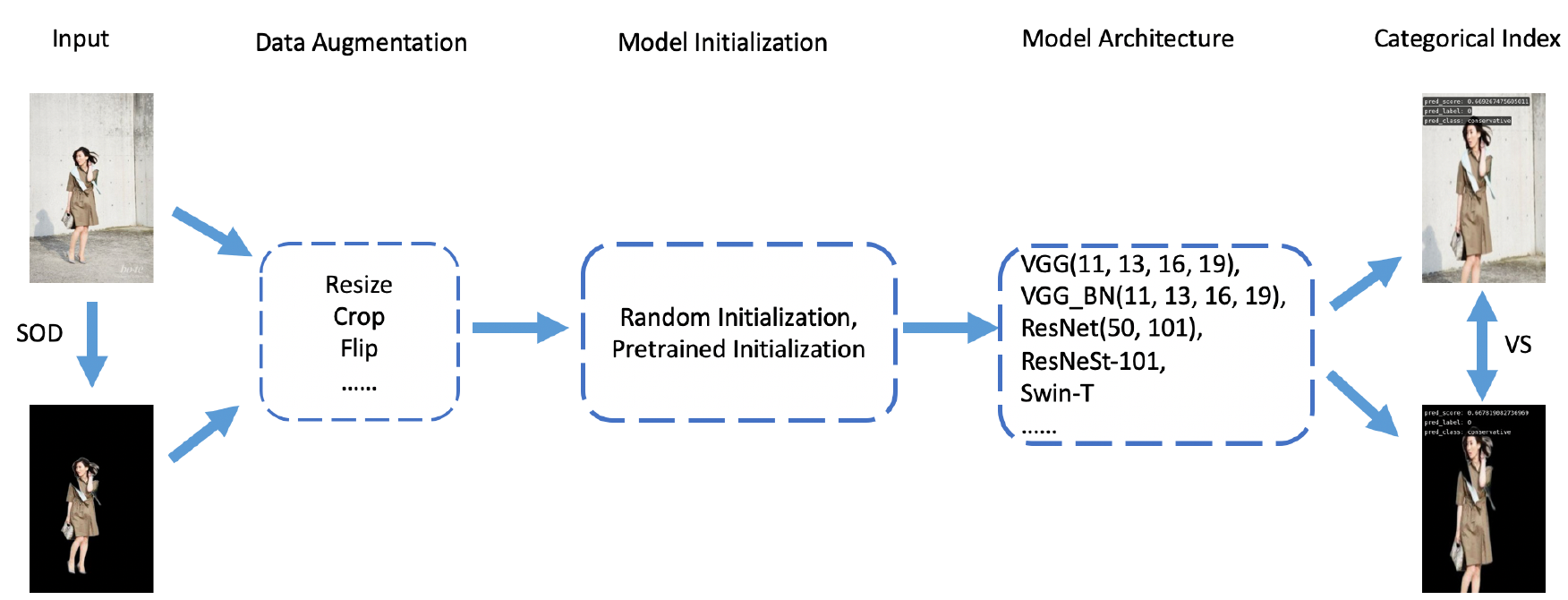}
    \caption{The pipeline for image classification experiments on FashionStyle14.}
    \label{fig:pipeline-of-classification}
\end{figure}

The pipeline for experiments to assess the impact of background removal in the fashion image classification task is shown in Figure~\ref{fig:pipeline-of-classification}. In this pipeline, we validate the value of background removal for different model architectures, model initialization, and data augmentation. We obtain the comparative results via multiple training settings and controlled inputs. The pipeline includes data augmentation at its preprocessing stage, model initialization, and selection at its model training stage. The input can be either original images or “rembg” images obtained by the SOD tool~ \cite{gatis_rembg_2022}, constituting a set of contrasting inputs. At the data augmentation step, the image augmentation operations can include resizing, cropping, flipping, or other compound policies like RandAugment~ \cite{cubuk_randaugment_2019}. At the step of model initialization, we compare the effect of background removal for two options: random initialization and pre-trained initialization. At a model architecture step, we compare the effect of background removal for several representative image classification models of various architectures and backbones, namely, the VGG series~\cite{simonyan_very_2015} with or without batch normalization~\cite{ioffe_batch_2015} layers, ResNet~\cite{he_deep_2015}, ResNeSt~\cite{zhang_resnest_2020}, and Swin Transformer~\cite{liu_swin_2021-1} \cite{liu_swin_2021}. The evaluation of the impact of background removal for models with different network depths, network backbones including CNN and Transformer, and with or without normalization can be accomplished by the combinations of these models. The output of classification is the categorical index of fashion style classes. Comparative evaluation 
experiments on the above 
zoo of various model architectures will help us comprehensively research the effect of background removal in fashion image classification.

Sincerely, an image classification model should be evaluated in multiple respects, such as quantitative accuracy, visual quality, inference speed, etc. However, our experiments focused on metrics for quantifying model accuracy rather than other aspects because we merely cared whether background removal of fashion data could improve model performance. Accuracy is essential for the classification task, which measures how positive and negative examples are correctly classified. As FashionStyle14 has only 14 classes, we consider only Top-1 accuracy, which measures the proportion of samples for which the predicted label matches the ground-truth label, i.e., an accurate prediction is a predicted class with the highest probability that is precisely the expected one. 

\subsection{Comparisons on Model Architecture}\label{subsec:comparison on architecture}
First, we evaluate the influence of several key attributes of a neural network, including network depth, normalization layer, block unit, backbone, and so on, on the classification performance of FashionStyle14 fashion images~\cite{takagi_what_2017} with removed backgrounds compared to the classification of the original FashionStyle14 images. Considering the interference of pre-trained model parameters, we train all the models from scratch using random parameter initialization. Below, we provide more details on our experiment setup. 

\textbf{Network depth.} We choose VGG11, VGG13, VGG16, and VGG19~\cite{simonyan_very_2015} as the representative models of shallow and simple networks; ResNet-50 and ResNet-101~\cite{he_deep_2015} standing for typical deep networks; ResNeSt-101~\cite{zhang_resnest_2020} as a supplementary to the deep networks with the attention architecture. We uniformly train the VGG series models for 100 epochs with a batch size of 32, with an SGD optimizer using a learning rate of 0.01 and a ”Step” decay learning rate scheduler. For ResNet-50 and ResNet-101, the batch size is 32, and the epoch is 100. Their optimizer is SGD with a learning rate of 0.1 and a ”Step” learning rate policy, which has minor changes from the original one trained on ImageNet~\cite{deng_imagenet_nodate}. ResNeSt is trained at a 16 batch size for 300 epochs, using an SGD optimizer with a learning rate of 6e-3 and a ”Cosine Annealing” learning rate policy~\cite{loshchilov_sgdr_2017} with a warm-up for a start.

\textbf{Normalization layer.} We add batch normalization~\cite{ioffe_batch_2015} layers based on the VGG series, which are usually named VGG11\_BN, VGG13\_BN, VGG16\_BN, and VGG19\_BN, forming the control group to the original VGG series. Batch normalization accelerates deep network training by reducing internal covariate shifts. We can also verify the value of normalization operation for training models to avoid background noise by comparing these results. As the experiments suggested, we found learning rate 5e-3 is better for the VGG series with batch normalization, and the rest training parameters keep the same as the VGG series. 

\textbf{Backbone Design.} In addition to traditional CNN architecture, we have also replenished the tiny version of the Swin Transformer~\cite{liu_swin_2021-1} (a.k.a Swin-T) represented for the transformer, which is currently popular model architecture. As our attempts indicated, the Swin-T obtained the best when we set batch size = 16, epoch = 400, using AdamW~\cite{loshchilov_decoupled_2019} optimizer with a learning rate of 1e-4 and a ”Cosine Annealing”~\cite{loshchilov_sgdr_2017} learning rate scheduler with a linear warm-up.

\begin{figure}[ht]
    \centering
    \includegraphics[width=1\columnwidth]{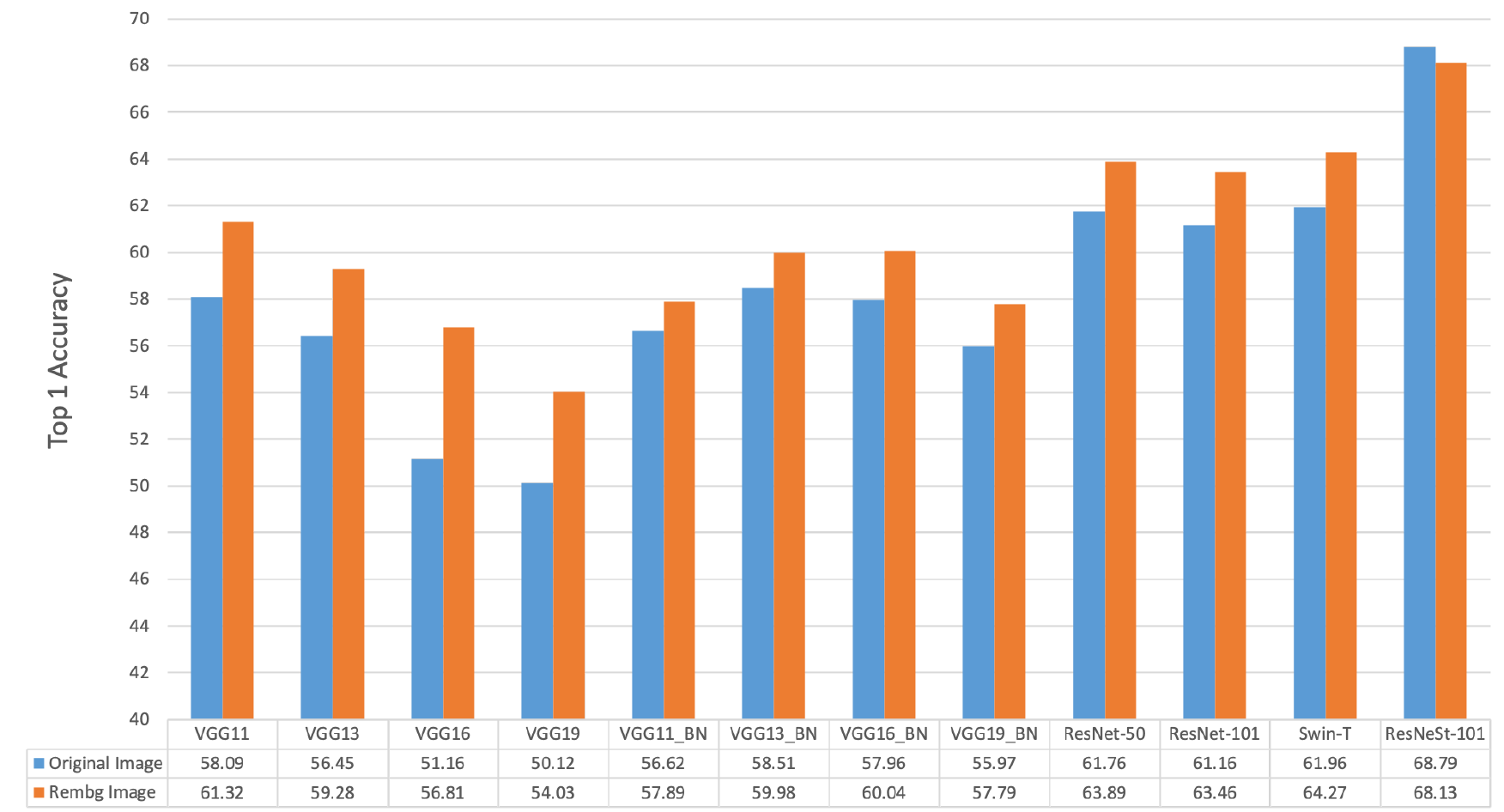}
    \caption{The performance of the trained from scratch models on rembg images with removed backgrounds compared to the original FashionStyle14 images.}
    \label{fig:comparisons-for-training-from-scratch}
\end{figure}

The more complex the model is, the more complicated its training converges, especially with limited resources for training models from scratch. We may see the performance of better models on ImageNet get worse results in the experiments. More importantly, we focus on the differences caused by the original and rembg images. The gap is usually visible even before the model reaches its best accuracy. Figure~\ref{fig:comparisons-for-training-from-scratch} compares the performance of the trained from scratch models on FashionStyle14 images with removed backgrounds (rembg) compared to the original images. As shown in Figure~\ref{fig:comparisons-for-training-from-scratch}, background-removed (rembg) images significantly improve performance in the VGG series, especially for VGG16 without batch normalization, which can improve over 5\% accuracy. The results meet our general expectation that background removal benefits model training by filtering the noise in advance, thus reducing learning difficulty.  

However, this improvement reduces a lot after adding batch normalization~\cite{ioffe_batch_2015}. The decline is due to a conflict between background removal and batch normalization. Batch normalization is a regularization technique to avoid overfitting due to insufficient quality labeled data. Similarly, data augmentation plays the same role by supplementing a dataset with similar data created from the original data in the dataset. On the contrary, instead of introducing random noises and transforming for better generalization, background removal removes lots of pixels that belong to the background, making regularization techniques more challenging to overcome overfitting. In other words, background removal plays a role opposite to batch normalization. Thus, batch normalization can not increase model accuracy with rembg images as much as with original images. From a cognitive point of view, background removal is equal to the human attention mechanism in the pre-processing step, leading the model to center on the foreground without interfering with the background. In contrast, batch normalization did the work during model training.

Meanwhile, as the network depth grows, the performance improvement from background removal still exists and is similar to the VGG\_BN series. This trend suggests that the model depth does not impact the performance improvements due to image background removal. Using rembg images is slightly worse than using the original image for the ResNeSt-101 that acquired the best accuracy. ResNeSt is featured by its Split-Attention Network, which can assist models in learning the foreground better. It suggests that background removal does not work for those models with attention mechanisms to distinguish foreground and background.

Besides, we extend the mean model accuracy to the accuracy per class (not shown for lack of space) and see that the effect of background removal involves all the classes rather than a specific class. We can confirm that background removal positively affects model training. In sum, background removal is a special kind of attention mechanism helping models to focus on key regions at the input layer of models while also increasing the risk of overfitting for deep networks due to the loss of the background information. 

On the one hand, for a simple and shallow network like VGG, the benefit of attention far outweighs the damage of regularization. On the other hand, with the growing complexity of deep networks and architectures, normalization like batch normalization and layer normalization is indispensable for models to enhance generalization and avoid overfitting caused by the shortage of enough labeled data. Thus, the advantages of background removal are inclined to decline until they disappear. Especially for the models with attention mechanisms, the effect of background removal is weaker or worse than the attention mechanism when the model is fully trained.

\subsection{Comparisons on Model Initialization}\label{subsec:comparison on initialization}
Model training from scratch is time-consuming and challenging to re-implement due to limited computer resources and random initialization. In most cases, one tends to use pre-trained model parameters on ImageNet as model initialization to fine-tune models swiftly. That is why we emphasize the value of background removal for the initialization from pre-trained models. In the following section, we compare the effect of background removal on the performance of the models with pre-trained initialization and those with random initialization. 

Pre-trained initialization is the first choice for most classification tasks as it improves the model convergence speed. The model parameters previously trained on a large dataset can be reused as the initial model parameters in the target domain, equal to training models with more data for better performance and generalization. To assess background removal benefits in image classification when training from scratch versus training from pre-trained initialization, we experiment with the VGG series [28], the VGG\_BN series~\cite{ioffe_batch_2015}, ResNet-50~\cite{he_deep_2015}, ResNet-101~\cite{he_deep_2015}, ResNeSt-101~\cite{zhang_resnest_2020}, and Swin-T~\cite{liu_swin_2021-1}. ViT-B~\cite{dosovitskiy_image_2021} is also considered supplementary to transformer architecture. 
VGG and VGG\_BN series retains the same training settings as those training from scratch, except for modifying the learning rate as 1e-4. For ResNet-50 and ResNet-101, we set a fixed learning rate of 1e-5, batch size of 32, and epoch 300. ResNeSt is fine-tuned for 200 epochs at a learning rate 5e-5 with a ”Step” policy, a batch size of 32. Swin-T and ViT-B are trained for 100 epochs at a fixed learning rate of 1e-5, with a batch size of 32. 

\begin{figure}[ht]
    \centering
    \includegraphics[width=1\columnwidth]{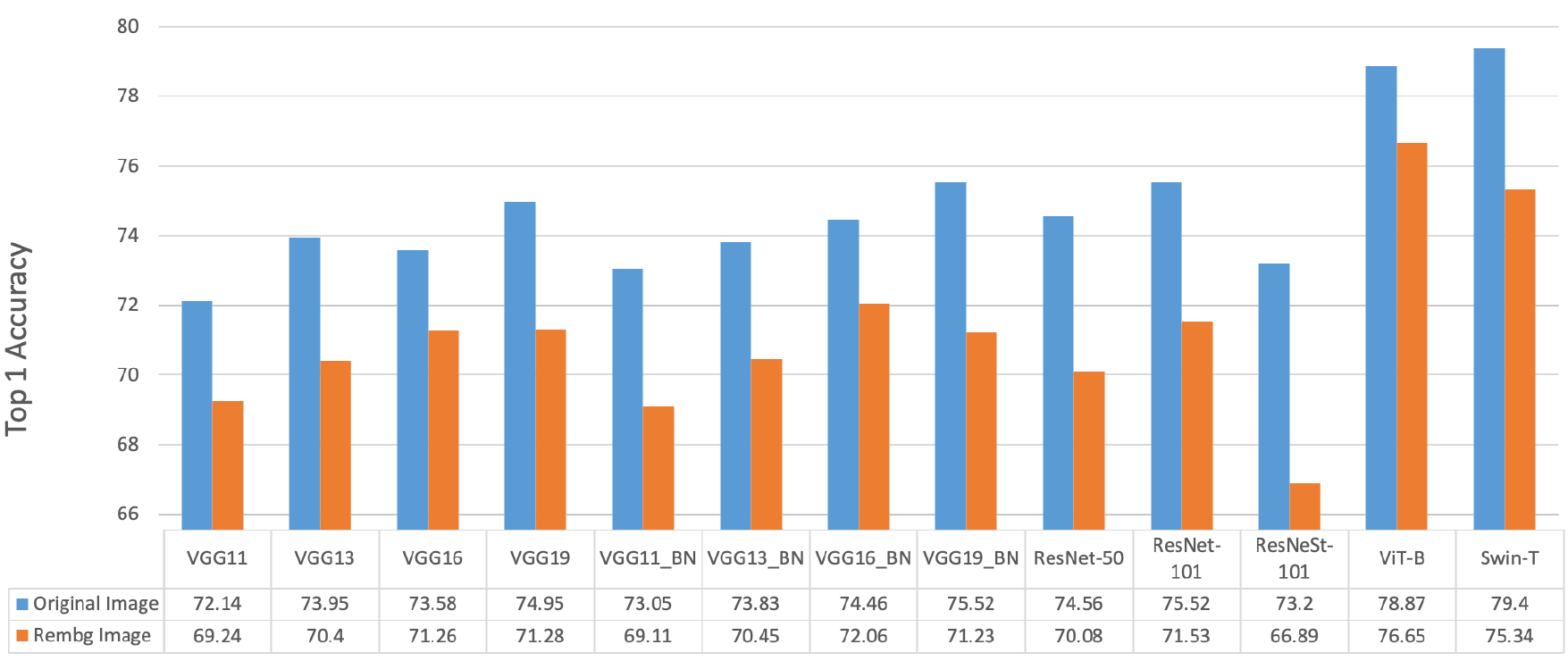}
    \caption{The performance of the fine-tuning pre-trained models on images with removed backgrounds (rembg) compared to the original ImageNet images.}
    \label{fig:comparisons-for-pre-trained-models}
\end{figure}

Figure~\ref{fig:comparisons-for-pre-trained-models} compares the performance of the fine-tuning pre-trained models on images with removed backgrounds (rembg) compared to the original ImageNet images. As we can see, all the models training from pre-trained parameters on ImageNet perform worse on rembg images than on original images. Combined with the results of Figure~\ref{fig:comparisons-for-training-from-scratch}, it is crystal clear that background removal is useless for fine-tuning models from pre-trained ones. This is because the original ImageNet images having diverse backgrounds are much different from the corresponding images without backgrounds, and these differences make transfer learning from ImageNet to fashion images without backgrounds more difficult. Another reason is related to model generalization. Pre-trained initialization and transfer learning enhance model generalization since the model can learn features with more data. In contrast, the model trained with rembg images is more susceptible to overfitting in deep networks. Besides, the models using pre-trained initialization perform better than the ones training from scratch. Thus, the performance improvement due to background removal is far less beneficial than using transfer learning by pre-trained initialization. 

In our more detailed evaluation of the per-class accuracy of models with pre-trained initialization (not shown for lack of space), we found that using rembg images decreases model accuracy in most categories. For ResNeSt-101, as shown in Figure~\ref{fig:comparisons-for-training-from-scratch}, it got no improvement with rembg images with random initialization, and it performs much worse with pre-trained initialization, as shown in Figure~\ref{fig:comparisons-for-pre-trained-models}. It manifests that background removal is impractical for models with attention mechanisms, no matter what kind of initialization.

\subsection{Comparisons on Data Augmentation}\label{subsec:comparison on data augmentation}

In addition to using pre-trained models for initialization, data augmentation is a valuable technique to boost model training. Image augmentations can be organized into two categories: spatial-level transformations, such as crop, flip, rotate, translate, shear, etc., and pixel-level transformations, such as brightness, sharpness, solarize, posterize, etc. 

\begin{figure}[ht]
    \centering
    \includegraphics[width=1\columnwidth]{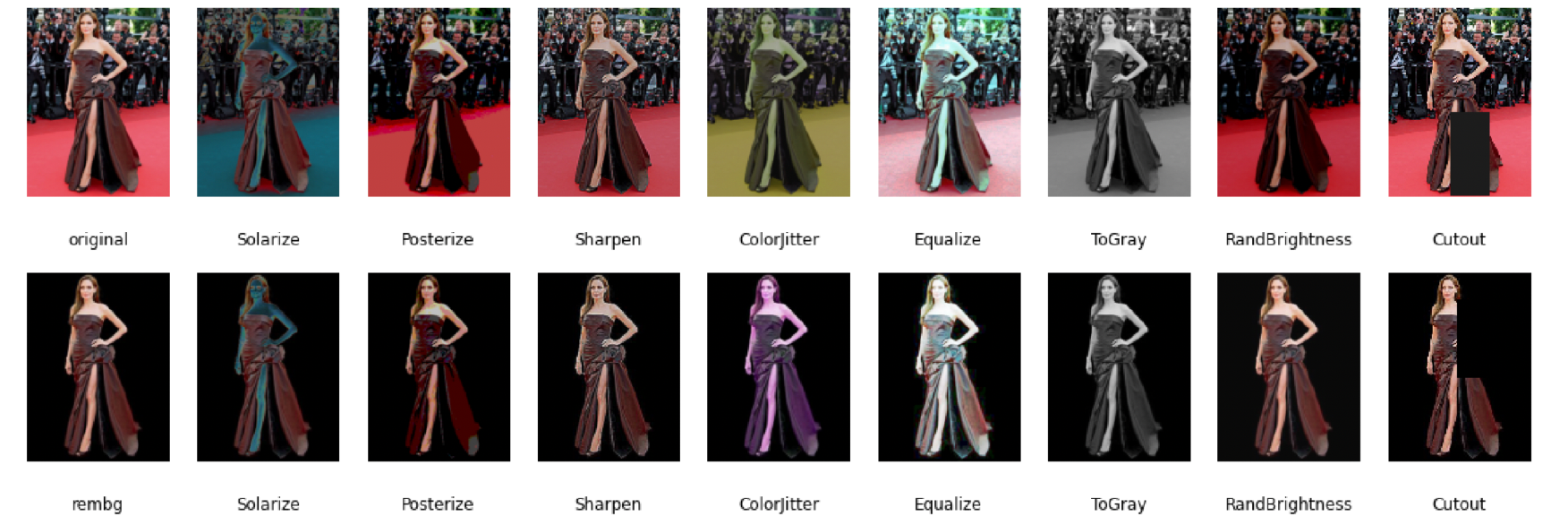}
    \caption{Samples of pixel-level transformations of original (first row) and rembg (second row) images. \cite{albumentations_2022}}
    \label{fig:pixel-transform}
\end{figure}

Theoretically, since we set the pixel value of the background to zero after removing the background, background pixels are no longer compatible with other data augmentation involved with pixel values. As shown in Figure~\ref{fig:pixel-transform}, we visually compare popular pixel-level transformers, including Solarize, Posterize, Sharpen, Colorjitter, Equalize, ToGray, Random Brightness, and Cutout~\cite{devries_improved_2017}, applied to the original and rembg images. These transformers applied to the whole image merely work on the pixels of the main object in the rembg image. However, as illustrated with Cutout, an image transformer can randomly erase parts of an image to improve model robustness; consequently, it fails to work with background removal owing to the lack of background. Similarly, background removal is incompatible with other regularization techniques like CutMix~\cite{yun_cutmix_2019} and MixUp~\cite{zhang_mixup_2018} that replace regions of images with other patches because they might mix up the blank pixel of the background with the original one. In other words, if we choose background removal as a preprocessing step, it becomes limited for model training to obtain improvement by currently existing data augmentation. It is demanding to introduce random noise for regularization and generalization when losing a large area of pixels that belong to the background.

Moreover, data augmentation is a technique of artificially extending a dataset with similar data created from the original dataset, targeted at releasing the overfitting of deep networks caused by the scarcity of enough annotated data. It is used to introduce randomness for better generalization, while background removal is used to simultaneously reduce the noise and information of images. There is a natural conflict between their measures and goals. The images applied with background removal do not perform well in the compatibility with other data augmentation. Furthermore, there is a risk of mixing up pure black clothes and existing items after setting the background pixels to zero. 

%% file: sections/image_segmentation.tex
In addition to the fashion image classification task, the image segmentation task is critical in fashion understanding with essential applications, especially for fine-grained fashion image classification. Thus, verifying the impact of background removal on the segmentation task is indispensable for our research on the effect of background removal in fashion image processing. Image segmentation can be formulated into three categories, classifying pixels with semantic labels (semantic segmentation), partitioning of individual objects (instance segmentation), or both (panoptic segmentation)~\cite{minaee_image_2020}. Semantic segmentation involves labeling each pixel in an image with an object category, resulting in a semantic image with categorical pixel values. Instance segmentation separates individual objects in an image, treating them as distinct entities, even if they belong to the same class. Essentially, instance segmentation expands on semantic segmentation by detecting and delineating each object region.

\begin{figure}[ht]
    \centering
    \includegraphics[width=1\columnwidth]{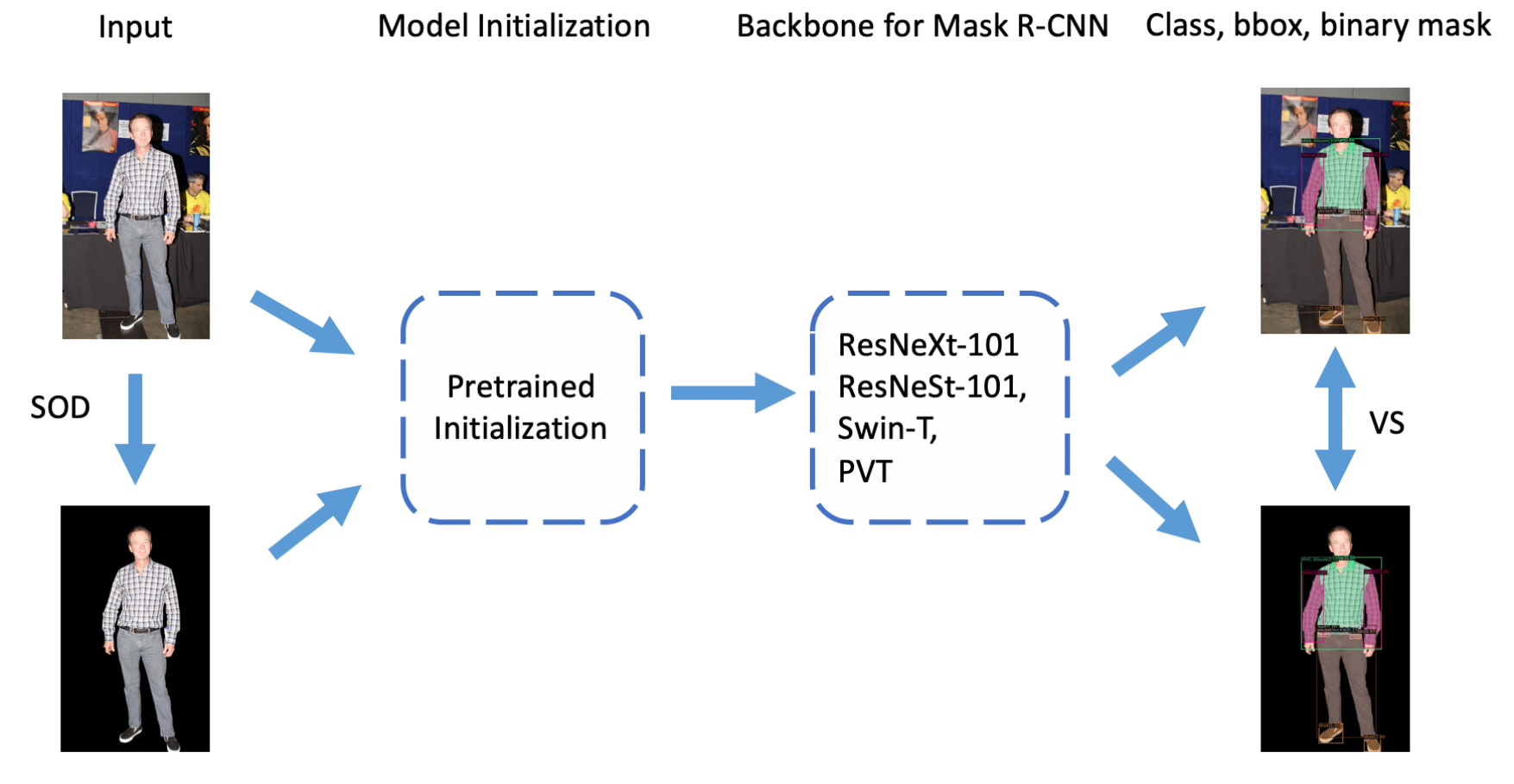}
    \caption{The pipeline of instance segmentation on Fashionpedia.}
    \label{fig:pipeline-of-instance-segmentation}
\end{figure}

\begin{figure}[ht]
    \centering
    \includegraphics[width=1\columnwidth]{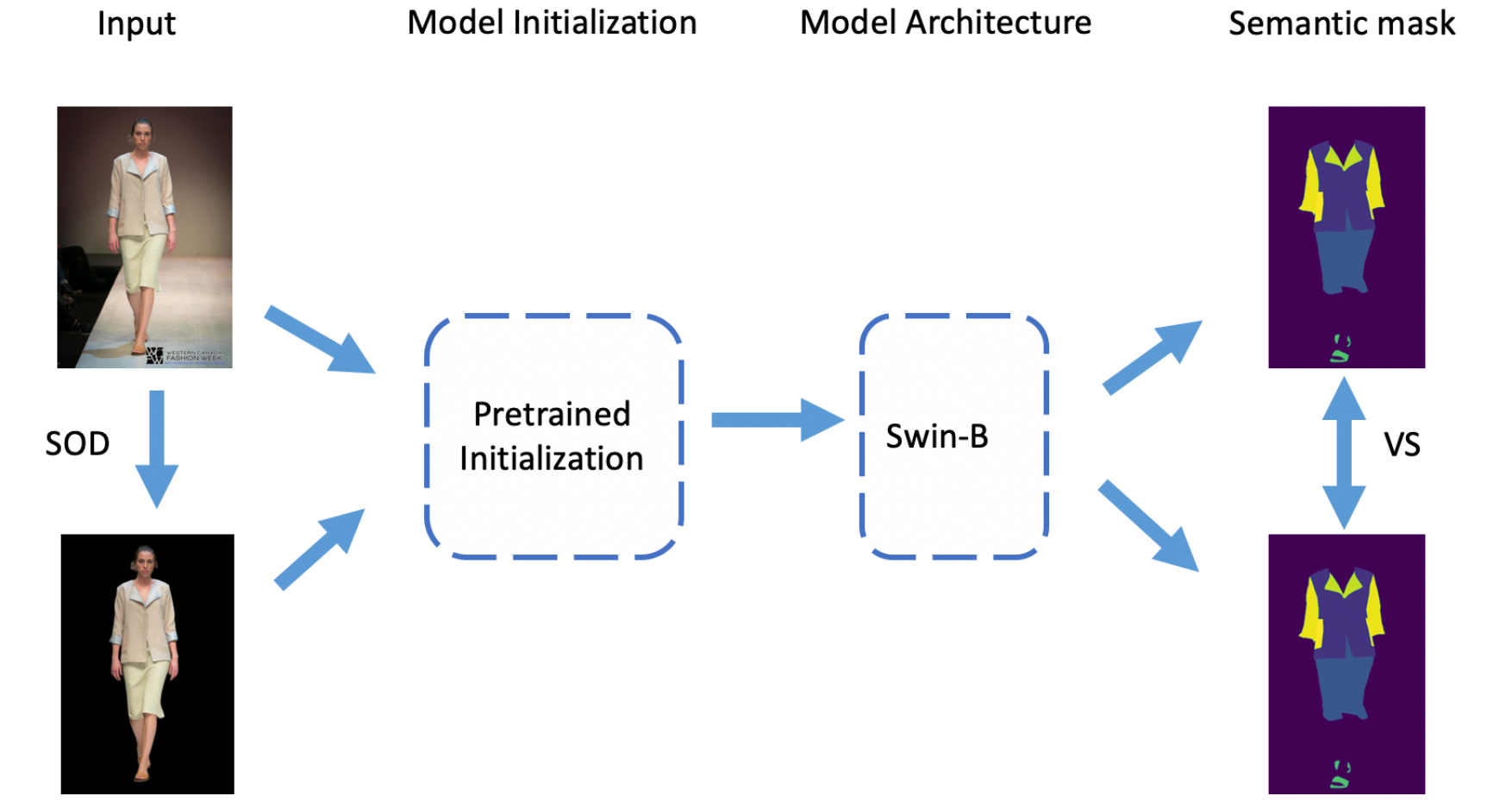}
    \caption{The pipeline of semantic segmentation on Fashionpedia.}
    \label{fig:pipeline-of-semantic-segmentation}
\end{figure}

After careful selection, we have chosen the Fashionpedia data set~\cite{jia_fashionpedia_2020} for the instance segmentation experiments and experiments on semantic segmentation with complete annotations. To assess the impact of background removal on fashion image segmentation, we designed two experiment pipelines for two kinds of image segmentation: instance segmentation and semantic segmentation, shown in Figure~\ref{fig:pipeline-of-instance-segmentation} and ~\ref{fig:pipeline-of-semantic-segmentation}, respectively. 

\subsection{Impact of Background Removal on Instance Segmentation.} 
The input to the pipeline for the experiments on instance segmentation (Figure~\ref{fig:pipeline-of-instance-segmentation}) is chosen between original images of Fashionpedia and corresponding \emph{rembg} images, with the annotations of class, locations of bounding boxes (bbox) and mask. To assess the impact of background removal on the performance of models for instance segmentation, we experiment with the models based on Mask R-CNN \cite{he_mask_2018} framework with various backbones to extract image features. Mask R-CNN is a highly influential model for instance segmentation. 

For our experiments with instance segmentation, we choose ResNeXt-101~\cite{xie_aggregated_2017} and ResNeSt-101~\cite{zhang_resnest_2020} represented for CNN, Swin-T~\cite{liu_swin_2021-1} and PVT~\cite{wang_pyramid_2021} represented for Transformer. The ResNeXt-101 and ResNeSt-101 are trained with a batch size of 4 due to the limitation of GPU memory. The optimizer is SGD with a learning rate of 0.02, and the ”Step” learning rate policy with a linear warm-up for 12 epochs. The training strategy for Swin-T and PVT uses AdamW~\cite{loshchilov_decoupled_2019} optimizer with a learning rate of 0.001 and a” Step” learning rate policy for 12 epochs. All the models utilize pre-trained ones on ImageNet as initialization, setting the input size as 512 x 512. There is no additional test for models trained from scratch since the cost of training segmentation models from scratch is relatively high and time-consuming, which is a thankless job and has less reference value. 

Table~\ref{tab:comparisons for instance segmentation} shows almost no difference between the performance of the Mask R-CNN with ResNeXt-101 or ResNeSt-101 backbone on original images and its performance on rembg images. Nevertheless, as presented in Section 4.3, those models trained by pre-trained initialization achieve much better with original images. We assert that under the Mask R-CNN architecture, RPN proposals are merely calculated by target bbox coordinates annotations, not affected by background removal. Thus, background removal does not impact mask generation, highly dependent on regions of interest and classification. Whereas using Swin-T and PVT as the backbone of Mask R-CNN, the results significantly drop when taking rembg images as input. The result is consistent with the previous experiments (Section~\ref{subsec:comparison on initialization}) for transformer models.

To sum up, background removal can not positively influence fashion image segmentation since the bbox and mask annotations can not be affected by background.

\begin{table}[h]
\centering
\caption{Comparisons w/o background removal for instance segmentation on Fashionpedia.}
\label{tab:comparisons for instance segmentation}
\resizebox{\columnwidth}{!}{  
\begin{tabular}{|l|ccc|ccc|}
\hline
 & \multicolumn{3}{c|}{bbox-mAP} & \multicolumn{3}{c|}{segm-mAP} \\ \cline{2-7} 
\multirow{-2}{*}{Backbone} & \multicolumn{1}{c|}{Original} & \multicolumn{1}{c|}{Rembg} & Diff. & \multicolumn{1}{c|}{Orignal} & \multicolumn{1}{c|}{Rembg} & Diff. \\ \hline
ResNeXt-101 & \multicolumn{1}{c|}{0.317} & \multicolumn{1}{c|}{0.316} & {\color[HTML]{3166FF} -0.001} & \multicolumn{1}{c|}{0.279} & \multicolumn{1}{c|}{0.277} & {\color[HTML]{3166FF} -0.002} \\ \hline
ResNeSt-101 & \multicolumn{1}{c|}{0.357} & \multicolumn{1}{c|}{0.354} & {\color[HTML]{3166FF} -0.003} & \multicolumn{1}{c|}{0.315} & \multicolumn{1}{c|}{0.313} & {\color[HTML]{3166FF} -0.002} \\ \hline
Swin-T & \multicolumn{1}{c|}{0.345} & \multicolumn{1}{c|}{0.325} & {\color[HTML]{3166FF} -0.020} & \multicolumn{1}{c|}{0.320} & \multicolumn{1}{c|}{0.299} & {\color[HTML]{3166FF} -0.020} \\ \hline
PVT & \multicolumn{1}{c|}{0.314} & \multicolumn{1}{c|}{0.306} & {\color[HTML]{3166FF} -0.009} & \multicolumn{1}{c|}{0.293} & \multicolumn{1}{c|}{0.278} & {\color[HTML]{3166FF} -0.015} \\ \hline
\end{tabular}
}
\end{table}

\subsection{Impact of Background Removal on Semantic Segmentation.} 

Semantic segmentation is also a common task for fashion image processing. The pipeline for semantic segmentation experiments is shown in Figure~\ref{fig:pipeline-of-semantic-segmentation}. The input is fashion images from Fashionpedia~\cite{jia_fashionpedia_2020}, with the semantic labels of ground truth generated by the annotation file, where the pixel value represents categories. The main difference from instance segmentation is that the output of semantic segmentation, also known as pixel-based classification, is a high-resolution semantic image with a categorical index (class) per pixel. For semantic segmentation, firstly, we need to generate semantic label annotation as ground truth, that is, a grey image of the same size as input with pixel value standing for categorical index. Semantic segmentation can be reckoned as binary classification per pixel. The background pixel can be ignored by setting a value of 255 since the categorical index starts at 0. Importantly, in this way, since the annotation of semantic segmentation can mark each background pixel, the background pixels will not be engaged in calculating the loss function. In simple words, the loss function of semantic segmentation is naturally supported for neglecting background by annotating pixel values of semantic labels. Thus, in theory, background removal should not work for semantic segmentation; therefore, we only experiment with Swin-B (Swin Transformer-Base)~\cite{liu_swin_2021-1} to validate the impact of background removal in the semantic segmentation task. 

The results of our evaluation experiments on background removal in semantic segmentation with Swin-B on fashion images from Fashionpedia are summarised in Table\ref{tab: comparisons for semantic segmentation}, which testify that background removal does not benefit semantic segmentation. The slight gap between the performance of Swin-B on original images and its performance on rembg images is caused by random initialization. 

\begin{table}[!h]
\centering
\caption{Comparisons w/o background removal for semantic segmentation on Fashionpedia}
\label{tab: comparisons for semantic segmentation}
\resizebox{\columnwidth}{!}{  
\begin{tabular}{|l|ccc|ccc|}
\hline
 & \multicolumn{3}{c|}{mIoU} & \multicolumn{3}{c|}{mAcc} \\ \cline{2-7} 
\multirow{-2}{*}{Model} & \multicolumn{1}{c|}{Original Img} & \multicolumn{1}{c|}{Rembg Img} & Diff. & \multicolumn{1}{c|}{Orignal Img} & \multicolumn{1}{c|}{Rembg Img} & Diff. \\ \hline
Swin-B & \multicolumn{1}{c|}{31.98\%} & \multicolumn{1}{c|}{31.74\%} & {\color[HTML]{3166FF} -0.24\%} & \multicolumn{1}{c|}{39.68\%} & \multicolumn{1}{c|}{39.43\%} & {\color[HTML]{3166FF} -0.25\%} \\ \hline
\end{tabular}}
\end{table}

Considering the classification, instance segmentation, and semantic segmentation, background removal is only beneficial in the classification task, especially when training simple models from scratch. Background removal cannot influence the bbox and mask location annotation of instance segmentation. The loss function of semantic segmentation is also approved to ignore background pixels. 

%% file: sections/conclusions.tex
We have empirically evaluated the impact of background removal on the performance of Neural Networks for fashion image classification and segmentation. We have experimented with various aspects of model training, including model architecture, initialization, data augmentation, and task type (classification and segmentation). Note that we limit our research to only fashion data and do not extend it to other data sources because our applied research has been primarily focused on the application domain of fashion image processing and understanding. Moreover, the background removal method we used, based on Salient Object Detection (SOD), has a certain limitation, as it can merely effectively remove background from the fashion images that usually have salient persons in clothes positioned in the center rather than the details or parts of fashion images. 

Our evaluation experiments show that, in most cases, background removal of fashion data is unnecessary in practice, even though, generally, a variety of backgrounds is one of the significant obstacles to training AI models to learn. With this perceptional premise, the models could be enhanced if removing diverse backgrounds. However, compared to the limited benefit of training shallow classification models from scratch, background removal has certain applicable limitations and conditions. Firstly, it is not sensible to fine-tune models from pre-trained ones. Besides, owing to the loss of background, background removal is not compatible with data augmentation techniques aiming to supplement the dataset by creating similar data. On one side, background removal can reduce noises and force models to learn key regions. On the other side, it adds the risk of overfitting for deep networks due to the loss of information and randomness. Similarly, regularization techniques like batch normalization are weakened by images without background. The positive effect of background removal is limited to simple and shallow networks that are not easy to suffer overfitting.

Nevertheless, our evaluation shows that background removal on fashion data benefits the fashion image classification task. In contrast, it provides no benefit for image segmentation, namely, instance segmentation and semantic segmentation with location annotation. Furthermore, background removal can facilitate model training to focus on the critical regions of interest only when the model cannot distinguish between foreground and background completely without adequate training strategy. However, background removal is mostly not an easy job for common data and images. Only fashion data, most images with full outfits visible, are relatively feasible to remove background by Salient Object Detection and related tools.